\documentclass{ieeeaccess}
\usepackage{cite}
\usepackage{amsmath,amssymb,amsfonts}
\usepackage{algorithmic}
\usepackage{graphicx}
\usepackage{textcomp}
\usepackage{bm}
\usepackage{cite}
\usepackage{fancyhdr}
\usepackage{hyperref}
\usepackage{multirow}
\makeatletter
\AtBeginDocument{\DeclareMathVersion{bold}
\SetSymbolFont{operators}{bold}{T1}{times}{b}{n}
\SetSymbolFont{NewLetters}{bold}{T1}{times}{b}{it}
\SetMathAlphabet{\mathrm}{bold}{T1}{times}{b}{n}
\SetMathAlphabet{\mathit}{bold}{T1}{times}{b}{it}
\SetMathAlphabet{\mathbf}{bold}{T1}{times}{b}{n}
\SetMathAlphabet{\mathtt}{bold}{OT1}{pcr}{b}{n}
\SetSymbolFont{symbols}{bold}{OMS}{cmsy}{b}{n}
\renewcommand\boldmath{\@nomath\boldmath\mathversion{bold}}}
\makeatother

\def\BibTeX{{\rm B\kern-.05em{\sc i\kern-.025em b}\kern-.08em
    T\kern-.1667em\lower.7ex\hbox{E}\kern-.125emX}}

\begin{document}
\history{Date of publication xxxx 00, 0000, date of current version xxxx 00, 0000.}
\doi{10.1109/ACCESS.2024.0429000}

\title{A Continual Learning Framework for Adaptive Control of Modular Soft Robots}
\author{\uppercase{Nilay Kushawaha}\authorrefmark{1,2} ,
\uppercase{Muhammad Sunny Nazeer}\authorrefmark{3}, ~\IEEEmembership{Member, IEEE,}, 
\uppercase{Baljinder Singh Bal}\authorrefmark{3,4}, \uppercase{Cecilia Laschi}\authorrefmark{3}, ~\IEEEmembership{Fellow, IEEE,}, and \uppercase{Egidio Falotico}\authorrefmark{1,2}, ~\IEEEmembership{Member, IEEE,}
}

\address[1]{The BioRobotics Institute, Scuola Superiore Sant’Anna, 56024, Pontedera, Italy}
\address[2]{Department of Excellence in Robotics and AI, Scuola Superiore Sant’Anna, 56127, Pisa, Italy}
\address[3]{Department of Mechanical Engineering, National University of Singapore, 119077, Singapore }
\address[4]{ETIS Lab UMR8051 CY Cergy Paris University - ENSEA - CNRS, France}
\tfootnote{We acknowledge the contribution from the Italian National Recovery and Resilience Plan (NRRP), M4C2, funded by the European Union–NextGenerationEU (Project IR0000011, CUP B51E22000150006, "EBRAINS-Italy"), PNRR-PRIN 2022 DISCOVER (CUP: J53D23002310006) as well as the support from the Italy-Singapore collaborative DESTRO project under A*STAR grant number R22I0IR124.\\
Github Repository Link : \url{https://github.com/nilay121/SMPL-A-Continual-Learning-Framework-for-Adaptive-Control-of-Modular-Soft-Robots}}


\corresp{Corresponding author: Nilay Kushawaha (e-mail: nilay.kushawaha@santannapisa.it).}

\begin{abstract}
Soft robots have attracted significant attention in applications such as medical intervention, rehabilitation, and robotic manipulation due to their inherent compliance, flexibility, and high degrees of freedom. Modular soft robots (MSRs), composed of multiple interconnected segments, represent an emerging class of robotic systems with highly deformable and reconfigurable structures capable of performing complex tasks. However, designing controllers for MSRs remains challenging due to their nonlinear dynamics, modeling complexity, and hyper-redundant nature. Existing approaches typically require controllers to be retrained from scratch whenever the robot morphology changes because of module attachment or detachment. Furthermore, for MSRs with fixed morphology, many methods rely on a single centralized controller trained across all modules, which limits modularity, reduces scalability, and can lead to error propagation across the system over time. In this work, we address these challenges through a continual learning inspired control framework capable of incrementally adapting to changes in robot morphology while preserving previously acquired knowledge. Specifically, the proposed framework enables the controller to sequentially learn new MSR configurations without forgetting previously learned ones. In addition, for MSRs with fixed configurations, the same framework can be employed in a distributed manner to learn module-specific dynamics, enabling localized control and improved precision. The proposed approach is validated through closed-loop trajectory tracking experiments in simulation using a tendon-driven soft robot, as well as on a real-world three-module pneumatic soft robotic arm. Furthermore, we demonstrate the adaptive capabilities of the framework through a reaching experiment in which the controller selectively activates only the necessary modules to reach a virtual target position, thereby reducing computational overhead.
\end{abstract}


\titlepgskip=-21pt

\maketitle

\section{Introduction}
Taking inspiration from the morphology of biological organisms, modular soft robots (MSRs) \cite{laschi2016soft} are designed to be flexible and compliant, enabling effective operation in unstructured environments \cite{rus2015design}. Due to inherent compliance and safe interaction with the environment, soft robots have found applications across diverse domains such as medical interventions, wearable technologies, navigation, and agriculture \cite{el2020soft,yasa2023overview}. The modular and reconfigurable design of MSRs offers enhanced flexibility and adaptability to diverse task requirements. Compared to single-module robots, MSRs possess a higher number of active degree of freedom (DOFs), allowing them to access larger workspaces and jointly perform more complex tasks. However, their hyper-redundancy, non-linear material properties, and complex dynamics make accurate modeling and control particularly challenging \cite{falotico2025learning}.
\begin{figure*}
	\centering
	\includegraphics[width=0.98\textwidth]{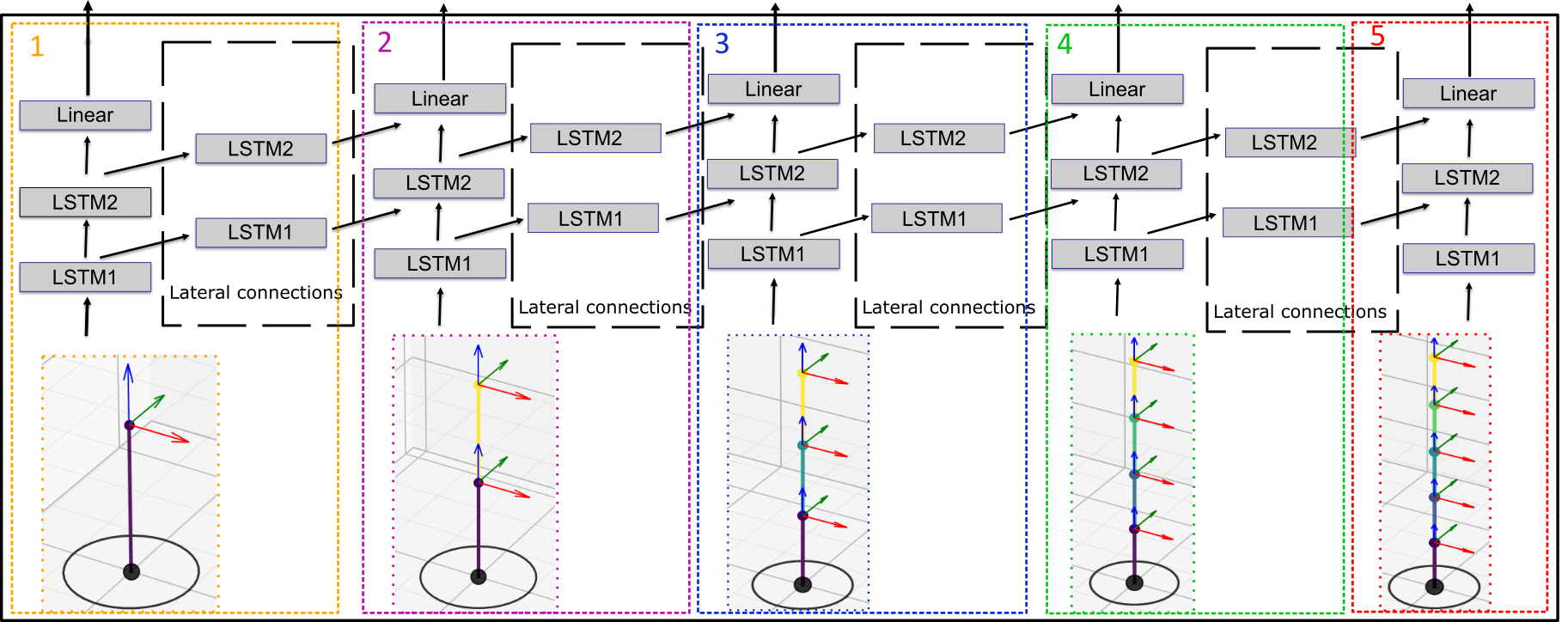}
	\caption{SMPL architecture for the Exp1S experiment. Sub-networks are incrementally added and trained on motor babbling data for different MSR configurations, with lateral connections enabling knowledge transfer during sequential learning. The framework is evaluated across five MSR configurations.}
	\label{fig:archi_Exp1s}
\end{figure*}

A wide range of modeling and control strategies have been proposed in the literature, broadly categorized into model-based and model-free approaches \cite{george2018control}. While model-based methods often provide faster and, in many cases, guaranteed convergence, their performance degrades when the robot operates in unpredictable environments \cite{nazeer2023policy} or when system dynamics evolve over time \cite{xie2025review}. Moreover, the non-linear and hysteretic behavior of soft materials, coupled with the increased dimensionality of input variables in modular configurations, further complicates the modeling process \cite{lipson2014challenges}. To overcome these limitations, a variety of learning-based approaches have been explored in the literature to capture the kinematics and dynamics of soft robots \cite{falotico2025learning}. Traditional machine learning (ML) methods have demonstrated an improved capability in handling non-linear dynamics compared to analytical techniques, particularly for MSRs \cite{johnson2021using}. Several studies have employed temporal neural architectures, such as long short-term memory (LSTM) networks \cite{thuruthel2019soft}, bi-directional LSTMs \cite{chen2024novel}, and echo state networks \cite{kuwabara2012timing}, to effectively model time-dependent and non-linear behaviors. In parallel, reinforcement learning (RL) approaches have been utilized to design controllers capable of adaptive motion planning \cite{nazeer2024rl}.

Despite these advances, most existing approaches assume a fixed robot morphology during training and typically require retraining from scratch whenever the robot configuration changes due to the addition or removal of modules. In practice, however, an effective MSR controller should be capable of incrementally incorporating information from newly added modules while retaining knowledge of previously learned configurations. Moreover, many existing methods for fixed-size MSRs rely on a centralized controller trained across all modules. Such approaches limit modularity, reduce scalability for robots with larger numbers of modules, and make the overall system more sensitive to error propagation across modules. To address these challenges, this work introduces a continual learning (CL)-based adaptive control framework \cite{kushawaha2024synapnet}.

\begin{figure*}
	\centering
	\includegraphics[width=0.98\textwidth]{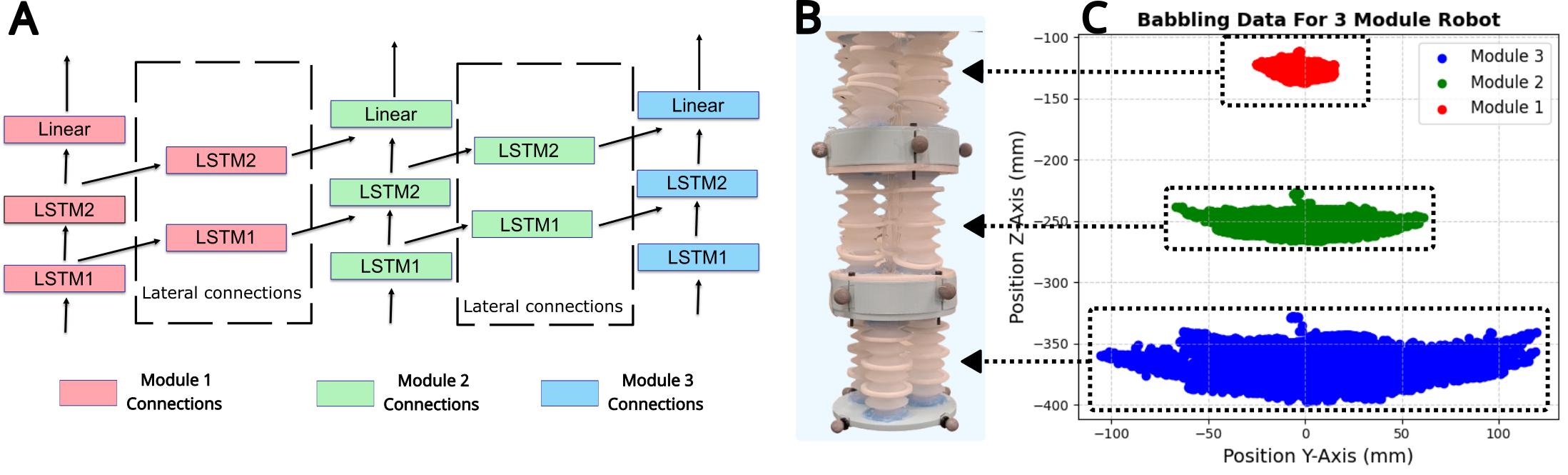}
	\caption{A: SMPL architecture for Exp2S and Exp2R, where each sub-network is trained on its corresponding MSR module; color coding matches the babbling data in (C), B: Three-module soft robot with markers for motion tracking, C: Workspace explored by each module during motor babbling for data collection.}
	\label{fig:archi_babb}
\end{figure*}
CL \cite{wang2024comprehensive}, also referred to as lifelong learning, addresses the challenge of learning from non-stationary data in an incremental manner, enabling the integration of new knowledge while preserving previously acquired information. In this work, we propose Soft Modular Progressive Learning (SMPL), a CL-based control framework designed to adapt to new MSR configurations as modules are attached or detached, as illustrated in Figure \ref{fig:archi_Exp1s}. Unlike conventional approaches, the proposed framework incrementally learns new configurations while retaining knowledge of previously encountered robot structures.

The SMPL framework models each MSR configuration using dedicated sub-networks while employing lateral connections to transfer relevant knowledge from previously learned configurations to newly introduced ones. In addition, we demonstrate the applicability of SMPL as a distributed control framework for MSRs with fixed morphology. In this setting, each sub-network is trained to locally control an individual module of the robot, as illustrated in Figure \ref{fig:archi_babb}. Such a distributed architecture enables finer and more interpretable control while also improving robustness in scenarios involving actuator degradation or failure (e.g., puncture), since only the affected module-specific sub-network requires retraining. The experimental evaluation presented in this paper is organized into three categories:

\begin{itemize}
\item Exp1S: A simulation experiment designed to evaluate the adaptive capabilities of the proposed SMPL framework under incremental changes in robot morphology through module attachment and detachment. 
\item Exp2S: A simulation study with a fixed MSR configuration, where each sub-network of the SMPL framework is trained to control the corresponding module of the MSR, enabling distributed control with improved precision and interpretability.
\item Exp2R: A real-world experiment analogous to Exp2S, evaluating the effectiveness of the proposed distributed control framework on a three-module soft robotic arm.
\end{itemize}

The primary contributions of this work are summarized as follows:

\begin{enumerate}
\item We propose a CL-based control framework capable of incrementally adapting to new MSR configurations as modules are attached, while preserving knowledge of previously learned configurations.

\item We demonstrate that the proposed framework can also learn module-specific dynamics for MSRs with fixed morphology, enabling distributed control with improved precision and scalability compared to conventional centralized approaches.

\item The proposed method is extensively evaluated through simulation experiments on a tendon-driven MSR and real-world experiments on a three-module pneumatic soft robotic arm across multiple trajectory tracking tasks.

\item To further demonstrate the practical applicability of the framework, we conduct an adaptive module activation experiment in which the robot selectively activates only the necessary modules to reach a virtual target position, thereby reducing computational overhead and improving control efficiency.
\end{enumerate}

The remainder of this paper is organized as follows: Section \ref{related_works} reviews related work on CL for modular architectures and MSR control strategies. Section \ref{methodology} details the proposed methodology. Section \ref{experiment_setup} describes the experimental setup for both simulation and real-world tests. Section \ref{results} presents the results of the trajectory tracking and adaptive module activation experiments. Finally, Section \ref{conclusion} concludes the paper by discussing key insights, limitations, and potential future directions.
 
\section{Related Works}
\label{related_works}
\subsection{Progressive Modular Architecture}
Modular network architectures consist of parallel sub-networks that enable incremental learning of new tasks while promoting knowledge sharing across components, thereby facilitating faster convergence. One of the most widely adopted modular approaches in CL is the Progressive Neural Network (PNN) \cite{rusu2016progressive}, which introduces a progressive architecture to enable effective knowledge transfer across sequential tasks. Several studies have explored the application of PNNs in both supervised learning and RL settings. 

For instance, \cite{juan2021shaping} combined a forward dynamics model with a PNN-based model-free RL framework to enable sim-to-real policy transfer. Similarly, \cite{tercan2018transfer} proposed a supervised PNN approach in which the initial sub-network is trained using extensive simulation data, followed by adaptation of subsequent sub-networks using limited real-world data. In another work, \cite{nguyen2022analytic} utilized a progressive modular framework to train an analytical, layer-wise deep neural network for modeling the kinematics and dynamics of an industrial robot. More recently, \cite{luo2024progressive} integrated PNNs within a model-based RL framework to address contact-rich in-hand manipulation tasks involving objects of varying sizes. Likewise, \cite{kushawaha2024synapnet,d2025semantization} proposed a brain-inspired modular architecture for object recognition using tactile sensory data acquired from a soft manipulator.

\subsection{Modular Soft Robot Control}
A variety of strategies have been proposed in the literature for controlling MSRs, which can generally be categorized into model-based and data-driven approaches. Model-based methods aim to analytically capture the kinematics of the system, often by deriving closed-form solutions under certain steady-state assumptions, as discussed in several works \cite{della2023model,best2021comparing}. In contrast, data-driven approaches rely on the latent representations present in the data to construct models of the robot without requiring complete prior knowledge of its morphology \cite{nk_cgan}, \cite{alessi2024pushing}. Data-driven methods can be further subdivided into supervised learning and RL techniques, each with its own pros and cons. 

\cite{chen2024novel} employed a bi-directional LSTM network incorporating explicit module-count information to design a controller for MSR in trajectory tracking tasks. Similarly, \cite{bianchi2024softsling} proposed two independent multi-layer perceptron (MLP) networks to achieve accurate object throwing into predefined targets using an MSR. In another work, \cite{pique2022controlling} introduced a CL-based framework to adapt the control of a soft robotic arm under varying load conditions. Reinforcement learning has also been widely explored; for instance, \cite{gan2022reinforcement} utilized RL to model the dynamics of a soft arm under unknown loads and interactions, enabling improved performance in contact-rich scenarios. More recently, \cite{nazeer2024rl} combined RL with imitation learning to develop a controller capable of high-precision reaching while addressing the sim-to-real transfer gap. Despite these advances, most existing approaches assume a fixed-length MSR during training and lack the capability for incremental adaptation as modules are added or removed. Furthermore, they typically rely on a single centralized model to control all modules, which limits scalability and modularity in control.

\section{Methodology}
\label{methodology}
\subsection{Problem Statement}
The objective of the proposed framework is to learn a mapping from the robot’s task space ($x$), defined by the end-effector's pose (position and orientation), to the actuation space($\tau$), thereby serving as a controller for the MSR. In the case of Exp1S, the framework is required to generate actuation commands for both the current module and all previously learned modules based on the end-effector pose. For instance, starting with a single-module MSR and subsequently adding an additional module, the SMPL framework should, after training on the new configuration, predict the actuation for both modules. Furthermore, when the second module is removed, the framework should retain the ability to generate actuation for the single-module configuration without additional retraining, demonstrating dynamic adaptability as the MSR structure evolves.
\begin{figure}
	\centering
	\includegraphics[]{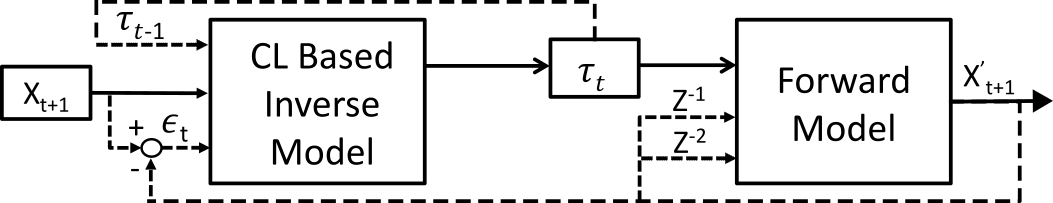}
	\caption{Closed-loop control architecture of the SMPL framework, comprising a CL-based inverse model and a forward dynamics model. The inverse model incorporates trajectory error as feedback for stable training, while the forward model is pretrained on motor babbling data and kept fixed during experiments.}
	\label{fig:control_architecture}
\end{figure}

For the second setting, corresponding to experiments Exp2S and Exp2R, the objective of the SMPL framework is to explicitly learn control policies for each module in an MSR with fixed morphology. In this scenario, the framework learns a mapping from the pose of each module to its respective actuation inputs. For example, for a three-module MSR, the SMPL architecture comprises three corresponding sub-networks, interconnected via lateral connections to facilitate knowledge transfer. Each sub-network is trained to capture the dynamics of an individual module, thereby encoding localized and module-specific information. As discussed previously, such a distributed control strategy is particularly advantageous when selective control of specific modules is required or when a module experiences degradation, as only the corresponding sub-network needs to be retrained.

To further enhance performance and ensure stability across all experimental settings, the SMPL framework is trained in a closed-loop configuration. Specifically, the actuation predicted by the SMPL-based inverse model is passed through a forward dynamics model, and the resulting error between the target trajectory and the predicted trajectory is fed back as an additional input to the inverse model, as illustrated in Figure \ref{fig:control_architecture}. This closed-loop formulation reduces task-space error and improves the robustness of the overall control architecture. For both cases, the forward and inverse models are trained in a supervised manner using babbling data.

\subsection{Progressive Neural Network}
One of the primary challenges in incremental learning is the degradation of performance on previously learned configurations after training on new ones, a phenomenon commonly referred to as catastrophic forgetting \cite{wang2024comprehensive}. PNNs address this issue through a structurally expandable architecture that preserves prior knowledge while enabling knowledge transfer via lateral connections. A PNN initially consists of a single sub-network with $L$ layers. When a new MSR configuration is introduced, as in Exp1S, an additional sub-network is instantiated. The parameters $\theta^{(1)}$ of the existing sub-network are frozen to retain previously acquired knowledge, while the parameters $\theta^{(2)}$ of the newly added sub-network are randomly initialized.

The hidden layers $h_i^{(2)}$ of the second sub-network receive inputs not only from their preceding layers within the same sub-network ($h_{i-1}^{(2)}$), but also from the corresponding layers ($h_{i-1}^{(1)}$) of the earlier sub-networks. This architectural design enables the integration of newly learned representations with previously acquired features, thereby facilitating forward knowledge transfer. The lateral connections regulate the flow of information from prior sub-networks to the current one, enhancing the model’s ability to generalize and adapt across configurations. A similar formulation is adopted for the Exp2S and Exp2R experiments; however, instead of associating each sub-network with a new MSR configuration, each sub-network is trained to capture module-specific dynamics for a fixed MSR structure.

The proposed SMPL framework leverages the PNN architecture to incrementally learn new MSR configurations or to encode module-specific information in fixed-size MSRs. Formally, the hidden representation of the $i$-th layer in the $k$-th sub-network is defined in Eq.~\ref{hidden_out}. The representation consists of two components: a \textit{self} contribution, obtained from the preceding layer of the current sub-network, and a \textit{lateral} contribution, transferred from the corresponding layers of previously learned sub-networks. Here, $k$ denotes the sub-network index and $i$ represents the hidden layer index. The term $W_i^{(k)}$ denotes the trainable weight matrix of the $i$-th layer in the $k$-th sub-network, $L_i^{(k:j)}$ represents the lateral connection weights between the current sub-network $k$ and the previous sub-networks $j$, and $f(\cdot)$ denotes a nonlinear activation function.

\begin{equation}
\label{hidden_out}
h_i^{(k)}(x) = f \left( \underbrace{W_i^{(k)}h_{i-1}^{(k)}(x)}_{\textit{self}} + \underbrace{\sum_{j<k}L_i^{(k:j)}h_{i-1}^{(j)}(x)}_{\textit{lateral}} \right)
\end{equation}

\subsection{Forward Model Description}
The forward model is designed to serve as an approximation of the robot’s forward dynamics during the training of the CL-based inverse model in a closed-loop configuration, where the end-effector error is fed back as an input to the inverse model. To this end, we adopt an LSTM-based architecture, owing to its proven effectiveness in modeling time-series data, as demonstrated in prior works \cite{thuruthel2019soft,pique2022controlling}. The forward model is trained in a supervised manner to learn the mapping from the actuation space to the task space of the robot. For Exp1S, a separate forward model is trained for each MSR configuration, whereas for Exp2S and Exp2R, individual forward models are trained for each module to better capture module-specific dynamics.

Specifically, the forward model takes as input the previous pose $x_{t-1}$, the current pose $x_{t}$, and the current actuation $\tau^{i}_{t}$, and predicts the pose at the next time step $x^{'}_{t+1}$. Here, $x = [p_x, p_y, p_z, \phi_x, \phi_y, \phi_z] \in \mathbb{R}^{6}$ represents the pose, corresponding to the end-effector in Exp1S and to individual modules in Exp2S and Exp2R. The terms $p_x, p_y, p_z$ denote position, while $\phi_x, \phi_y, \phi_z$ represent orientation, expressed as the third column of the rotation matrix corresponding to the local $z$-axis in the world frame. The actuation $\tau^{i} \in \mathbb{R}^{2}$ for each module in simulation experiments, whereas for real-world experiments (Exp2R) it lies in $\mathbb{R}^{3}$ for the second and third modules and in $\mathbb{R}^{4}$ for the first module. Once trained, the forward model parameters remain fixed during both training and evaluation of the CL-based inverse model across all experiments. The mathematical formulation for the $i$-th module is given in equation \ref{fwd_eqn}.

\begin{equation}
    \label{fwd_eqn}
    x^{'i}_{t+1} = f_{fwd}\big( x^{i}_{t-1}, x^{i}_t, {\{\tau^{i}_t\}}^{i}_{j=0} \big)
\end{equation}

In the case of Exp2S and Exp2R, the forward model for the first module depends solely on its own actuation, whereas the forward models for subsequent modules additionally incorporate the actuation signals of preceding modules. This design choice accounts for inter-module coupling, and empirical observations indicate that including actuation from earlier modules significantly reduces task-space prediction error. Similarly, in Exp1S, the forward model considers the actuation inputs of all modules corresponding to the given MSR configuration.

The LSTM architecture consists of two hidden layers with 64 neurons each, followed by a linear output layer. A sequence length of 15 was empirically found to accelerate convergence and reduce prediction error during training. The model is trained using a batch size of 64, a learning rate of $1 \times 10^{-3}$, and optimized over 800 epochs per module using a mean squared error (MSE) loss function.

\subsection{SMPL Training}
\label{SMPL_training_labe}
The training procedure of the SMPL framework is consistent across Exp1S, Exp2S, and Exp2R, with the primary distinction being the specific outputs learned by each sub-network in the respective settings. For Exp1S, as illustrated in Figure \ref{fig:archi_Exp1s}, training begins with a single sub-network that learns the inverse dynamics of a single-module MSR using supervised learning on motor-babbling data. After the first sub-network has converged, a second sub-network is introduced, and lateral connections between the two sub-networks are initialized with random weights. The parameters of the first sub-network are subsequently frozen to mitigate catastrophic forgetting, while the second sub-network is trained using motor-babbling data corresponding to the two-module MSR configuration. During this phase, the same inputs are also propagated through the first sub-network, enabling knowledge transfer via lateral connections. This process can be naturally extended to an $N$-module MSR, where a dedicated sub-network is incrementally trained for each new configuration, resulting in a scalable and dynamically adaptive learning framework.

In the case of Exp2S and Exp2R, as depicted in Figure \ref{fig:archi_babb}(A), each sub-network is associated with a specific module of a fixed three-module MSR, as shown in Figure \ref{fig:archi_babb}(B). Here, each sub-network is trained independently using motor-babbling data corresponding to its respective module, thereby enabling more precise and localized control.

Across all experiments, each sub-network receives as input the target pose $x_{t+1}$, the current task-space error $\epsilon_t$, and the previous actuation $\tau_{t-1}$, and outputs the actuation $\tau_t$ required to reach the desired pose, as illustrated in Figure \ref{fig:control_architecture}. At initialization, both the previous actuation $\tau_{t-1}$ and the pose error $\epsilon_{t}$ are set to zero; these values are subsequently updated as the inverse model generates predictions. In Exp1S and Exp2S, the pose error is computed by passing the predicted actuation through the forward model, whereas in Exp2R it is obtained directly from real-world robot observations. This procedure is applied during both training and inference. The formulation is given by:

\begin{equation}
\tau_t = f_{inv}\big(x_{t+1}, \epsilon_t, \tau_{t-1} \big)
\end{equation}
\begin{equation}
\epsilon_t = \left|x_{t+1} - x^{'}_{t+1}\right|
\end{equation}

Each column of the PNN architecture consists of two LSTM layers with 32 hidden units each, followed by a linear output layer. Similar to the forward model, a sequence length of 15 is employed for the LSTM inputs. The network is trained using a batch size of 32, a learning rate of $1 \times 10^{-3}$, and optimized for 800 epochs using the Adam optimizer.

\section{Experimental Setup}
\label{experiment_setup}
\begin{figure*}
	\centering
	\includegraphics[]{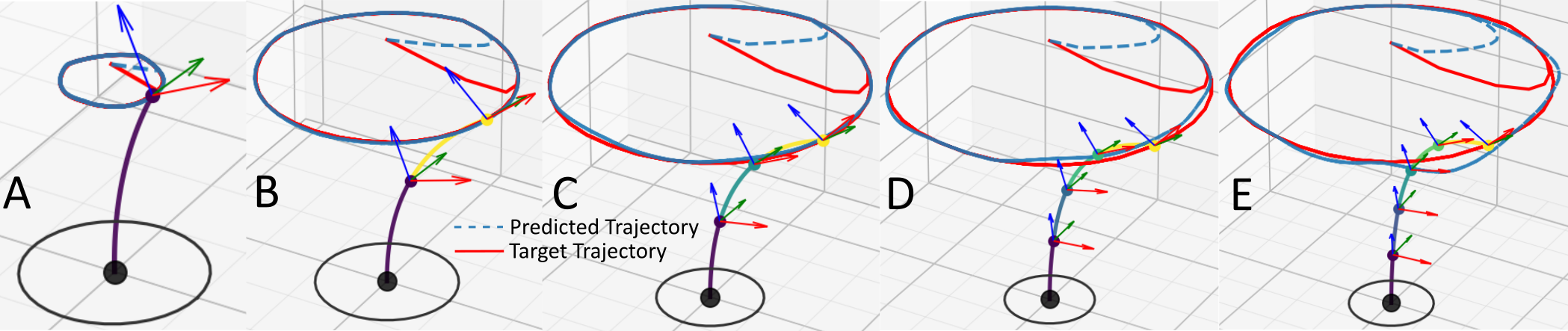}
	\caption{Figures A–E illustrate the incremental addition of modules to the MSR and the SMPL algorithm’s ability to retain knowledge across configurations. After training on the five-module setup (E), performance is evaluated on all prior configurations. The solid red line denotes the target trajectory, and the blue dashed line represents SMPL predictions.}
	\label{fig:circle_exp1s_all_mods}
\end{figure*}
\subsection{Simulation Experiment}
We first evaluate the proposed SMPL framework in a simulation environment, following the setup described in \cite{chen2024novel}. The simulator models a cable-driven soft robot using the PyElastica framework \cite{zhang2019modeling}, based on the Cosserat rod theory \cite{alqumsan2019robust}. Each module has a length of 50 mm and is independently actuated using four cable-driven actuators operating at 1 Hz. During motor babbling in simulation, two of the four actuation signals are generated randomly, while the remaining two are defined as their negatives to maintain actuation symmetry. In the case of Exp1S, we record the end-effector pose of the MSR across five different configurations, as illustrated in Figure \ref{fig:archi_Exp1s}, along with the corresponding actuation signals generated during motor babbling.

For the remaining experiments (Exp2S and Exp2R), we record both the pose and corresponding actuation of each module for a fixed three-module MSR, in both simulation and real-world settings. For additional details on the simulation environment, readers are referred to \cite{chen2024novel} or the accompanying github repository. For training, we collect 14,000 motor babbling samples generated through random actuation to ensure sufficient exploration of the workspace. During data collection, all modules remain active to ensure comprehensive coverage of the system dynamics. From each data sample, position and orientation vectors are extracted to construct a structured dataset comprising actuation inputs and corresponding pose outputs. In addition to the babbling dataset, we also collect 1,000 additional samples each for three custom trajectories: a circular shape, a rectangular shape, and a spiral trajectory to assess the performance of the controllers on some arbitrary shapes. All actuation signals and pose values are normalized within the range $[-1, 1]$ prior to being used by the respective sub-networks to ensure faster convergence.

\subsection{Real Robot Experiment}
A three-module soft robotic arm is employed where the proximal module (connected with the base) is composed of four linear bellow actuators spaced $90^\circ$ apart while the subsequent two modules are composed of three linear bellow actuators each spaced $120^\circ$ apart. The four actuators in the proximal module allows it to carry the weight of the subsequent modules while maintaining actuation. Each actuator is 80mm long with a 30mm outer and 25mm inner diameter. The actuator design features 7 equidistant convolutions with 1mm pitch. Each convolution comprises of a bulging structure and a deep valley; The wall thickness is higher at the bulge and thinner at the deep valley, allowing axial deformation while limiting radial expansion, enabling purely linear motion (extension under positive pressure and contraction under negative pressure). Actuators can achieve up to $~120\%$ elongation at $90~KPa$ and up to $70\%$ contraction at $-40KPa$ relative to their characteristic length. Combining multiple linear actuators together enables the soft arm to achieve isotropic movements including omni-directional bending. The actuators are fabricated by injection molding using silicon with 30A shore hardness. The design and characterization details can be found in our previous work~\cite{NazeerBellows2025}.

As in the simulation study, we begin by collecting 14,000 motor babbling samples generated by sending random pressure actuation to the MSR while keeping all the modules active. The precise motion of each module is recorded using a motion capture system (OptiTrack PrimeX13), which tracks reflective markers placed on the robot. Specifically, each module is equipped with four reflective markers, and an additional four markers are fixed at the top of the robot to serve as reference points. From the motion capture system, we extract the positions and the orientation vector for each module, and construct a structured dataset as explained before. Figure \ref{fig:archi_babb} (C) shows the workspace covered by the three modules of the MSR during the motor babbling phase. Similar to the simulation setup, we also collect additional data for the custom shape trajectories. For all the experiments the robot is actuated at 5 Hz, and both actuation as well as position values are normalized to $[-1, 1]$ range.

\begin{table*}[t]
\caption{Pose error of the proposed SMPL algorithm across different MSR configurations after incremental training (Exp1S). Results are reported for both closed-loop and open-loop settings over multiple trajectory tracking tasks.}
\label{tab_exp1s_error}
\centering
\footnotesize
\renewcommand{\arraystretch}{1.2}

\resizebox{500pt}{!}{
\begin{tabular}{|c|c|c|c|c|c|c|c|c|c|c|c|}
\hline

\multirow{2}{*}{\shortstack{Trajectory \\ Type}} & \multirow{2}{*}{Metrics}  
& \multicolumn{5}{c|}{$\#$ of Modules (Closed Loop)} 
& \multicolumn{5}{c|}{$\#$ of Modules (Open Loop)} \\

\cline{3-12}
& & 1 & 2 & 3 & 4 & 5 & 1 & 2 & 3 & 4 & 5 \\
\hline

\multirow{2}{*}{\shortstack{Test \\ Babbling}} 
& Pos (mm)  & \textbf{0.16}$_{\pm0.03}$ & \textbf{0.60}$_{\pm0.23}$ & \textbf{2.91}$_{\pm1.33}$ & \textbf{5.27}$_{\pm2.96}$ & \textbf{6.89}$_{\pm3.25}$ & 0.17$_{\pm0.07}$ & 0.71$_{\pm0.39}$ & 3.55$_{\pm2.61}$ & 5.44$_{\pm3.23}$ & 7.38$_{\pm3.45}$ \\
& Ori (deg) & \textbf{0.38}$_{\pm0.08}$ & \textbf{0.69}$_{\pm0.17}$ & \textbf{2.40}$_{\pm2.03}$ & \textbf{3.90}$_{\pm1.57}$ & \textbf{5.32}$_{\pm2.11}$ & 0.40$_{\pm0.23}$ & 0.79$_{\pm0.42}$ & 2.94$_{\pm1.73}$ & 4.12$_{\pm2.76}$ & 5.56$_{\pm2.91}$ \\ 
\hline

\multirow{2}{*}{Circle} 
& Pos (mm)  & \textbf{0.15}$_{\pm0.05}$ & \textbf{0.48}$_{\pm0.22}$ & \textbf{1.75}$_{\pm0.85}$ & \textbf{3.02}$_{\pm1.93}$ & \textbf{4.00}$_{\pm2.32}$ & 0.16$_{\pm0.12}$ & 0.56$_{\pm0.26}$ & 1.89$_{\pm0.95}$ & 3.03$_{\pm2.01}$ & 4.31$_{\pm3.11}$  \\
& Ori (deg) & \textbf{0.37}$_{\pm0.10}$ & 0.58$_{\pm0.31}$ & \textbf{1.09}$_{\pm0.69}$ & \textbf{1.48}$_{\pm0.87}$ & \textbf{2.58}$_{\pm1.95}$ & 0.38$_{\pm0.17}$ & \textbf{0.57}$_{\pm0.34}$ & 1.55$_{\pm1.33}$ & 2.08$_{\pm1.97}$ & 2.82$_{\pm1.63}$ \\ 
\hline

\multirow{2}{*}{Rectangle} 
& Pos (mm)  & \textbf{0.13}$_{\pm0.08}$ & \textbf{0.30}$_{\pm0.15}$ & \textbf{1.39}$_{\pm0.98}$ & \textbf{2.65}$_{\pm1.32}$ & \textbf{2.71}$_{\pm1.69}$ & 0.16$_{\pm0.04}$ & 0.69$_{\pm0.41}$ & 1.75$_{\pm1.52}$ & 2.67$_{\pm1.88}$ & 3.02$_{\pm2.13}$  \\
& Ori (deg) & \textbf{0.31}$_{\pm0.19}$ & \textbf{0.42}$_{\pm0.31}$ & \textbf{1.03}$_{\pm0.75}$ & \textbf{1.42}$_{\pm0.86}$ & 2.03$_{\pm1.28}$ & 0.33$_{\pm0.27}$ & 0.67$_{\pm0.29}$ & 1.48$_{\pm1.02}$ & 1.93$_{\pm1.15}$ & \textbf{1.82}$_{\pm1.49}$ \\
\hline

\multirow{2}{*}{Spiral} 
& Pos (mm) & \textbf{0.12}$_{\pm0.05}$ & \textbf{0.39}$_{\pm0.24}$ & \textbf{1.49}$_{\pm0.82}$ & \textbf{2.32}$_{\pm1.02}$ & \textbf{2.15}$_{\pm1.37}$ & 0.56$_{\pm0.30}$ & 0.73$_{\pm0.52}$ & 2.11$_{\pm1.59}$ & 2.36$_{\pm1.72}$ & 2.66$_{\pm1.80}$  \\
& Ori (deg) & \textbf{0.29}$_{\pm0.11}$ & \textbf{0.52}$_{\pm0.37}$ & \textbf{1.12}$_{\pm1.03}$ & \textbf{1.48}$_{\pm1.10}$ & \textbf{1.88}$_{\pm0.78}$ & 0.36$_{\pm0.17}$ & 0.70$_{\pm0.15}$ & 1.94$_{\pm1.18}$ & 1.85$_{\pm1.02}$ & 1.95$_{\pm0.96}$ \\
\hline

\end{tabular}
}

\end{table*}

\subsection{Evaluation Metrics \& Benchmarking}
The collected motor babbling dataset is partitioned into training $(70\%)$, validation $(10\%)$, and test $(20\%)$ subsets. The proposed method and all benchmark models are evaluated on the test babbling dataset comprising 2,800 samples, as well as on the three predefined trajectories introduced earlier, each consisting of 1,000 samples. For all the experiments, we report the end-effector position error (equation ~\ref{pos_error}), computed as the L2-norm between the target and predicted positions for a given trajectory. The orientation error is measured in degrees using equation \ref{cosine_error}, where $\phi_p$ and $\phi_t$ denote the predicted and target orientation vectors, respectively.

\begin{figure}
	\centering
	\includegraphics[]{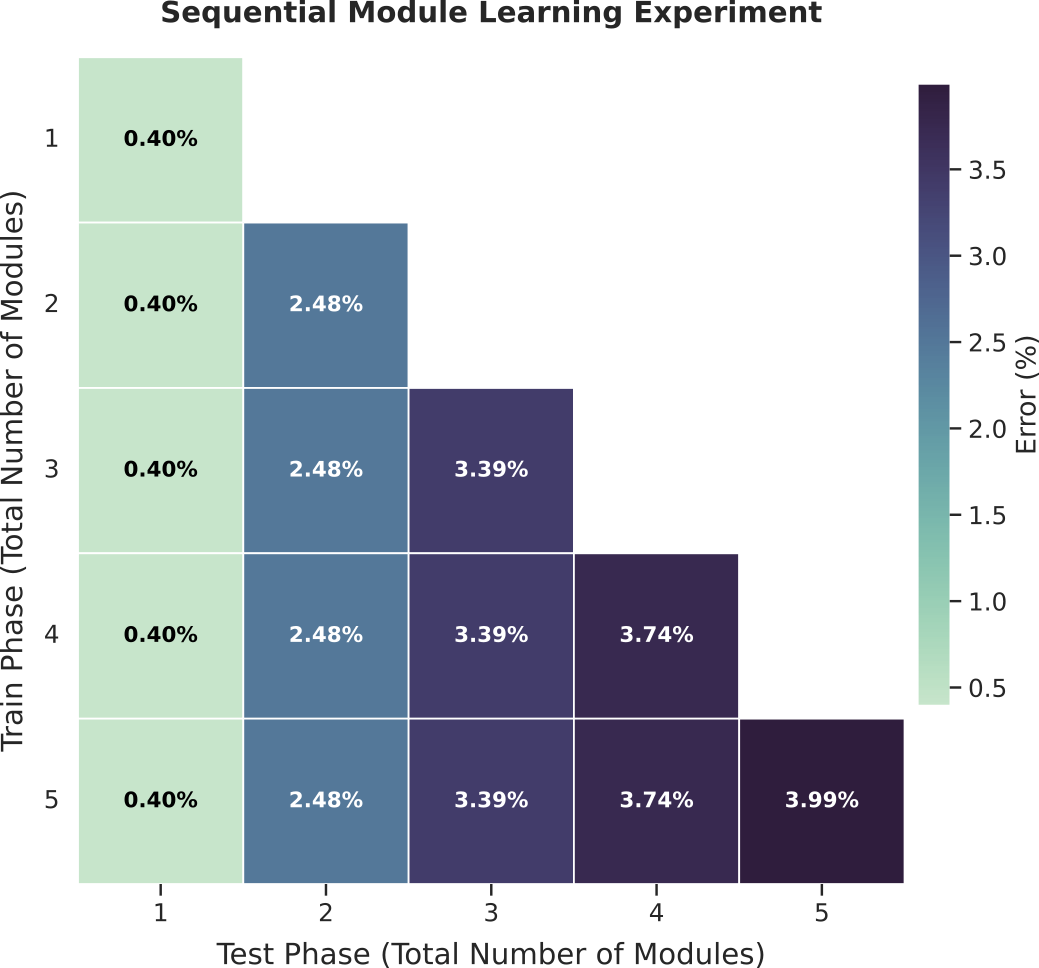}
	\caption{Percentage normalized tip error $(E^{L}_{\text{tip}})$ for the Exp1S experiment after incremental training on each MSR configuration. After each stage, the SMPL algorithm is evaluated on all previously learned configurations.}
	\label{fig:cf_msr}
\end{figure}

Additionally, for Exp1S, we also report the percentage normalized tip error, defined as the ratio of the end-effector position error to the total length of the MSR, as given in equation \ref{relative_error}, where $L$ represents the total robot length for a specific configuration. Position error is expressed in millimeters (mm), orientation error in degrees, and normalized tip error in percentage. For the Exp2S and Exp2R experiments, both position and orientation errors are averaged across all three modules of the MSR.

\begin{equation}
    \label{pos_error}
    X_{error} =  \left \| X_p - X_t \right\|_2
\end{equation}
\begin{equation}
    \label{cosine_error}
    \phi_{error} = \arccos\big({\frac{\phi_p\cdot\phi_t}{\left\|\phi_p\right\|\left\|\phi_t\right\|}\big)}
\end{equation}
\begin{equation}
    \label{relative_error}
    E^{L}_{\text{tip}} (\%) = \frac{X_{\text{error}}}{L} \times 100
\end{equation}

To benchmark the proposed approach, four baseline models are implemented. These include a bi-directional LSTM following \cite{chen2024novel}, a standard LSTM as in \cite{wang2023soft}, and a variational autoencoder (VAE)-LSTM model trained using variational inference as described in \cite{yoshimitsu2023forward}. In addition, a three-layer multi-layer perceptron (MLP) is implemented to provide an approximate upper bound on achievable error. All models are trained within a closed-loop configuration, and hyperparameters are optimized using the Optuna framework \cite{akiba2019optuna} based on validation performance. Each algorithm is evaluated over five independent trials with different random seeds, and the mean along with standard deviation is reported for all metrics.

\section{Results \& Discussion}
\label{results}
\subsection{Trajectory Tracking Experiment}
We first evaluate the performance of the proposed framework in the Exp1S setting. In this scenario, the model is trained incrementally across different MSR configurations, starting from a single-module robot and extending up to five modules. Figure \ref{fig:circle_exp1s_all_mods} (A–E) demonstrates the ability of the proposed approach to generate circular trajectories for all previously encountered configurations after sequential training. The SMPL framework achieves accurate tracking across all five configurations, with minor deviations observed in later configurations, likely due to the increased size and complexity of the robot. Additionally, the normalized tip error $(E^{L}_{\text{tip}})$ obtained after training on each configuration is illustrated in Figure \ref{fig:cf_msr}. The error is calculated on the "test babbling" dataset after each training stage. The results indicate that the model successfully acquires control over newly introduced configurations while retaining performance on previously learned ones, demonstrating its ability to mitigate catastrophic forgetting.

\begin{figure}
	\centering
	\includegraphics[]{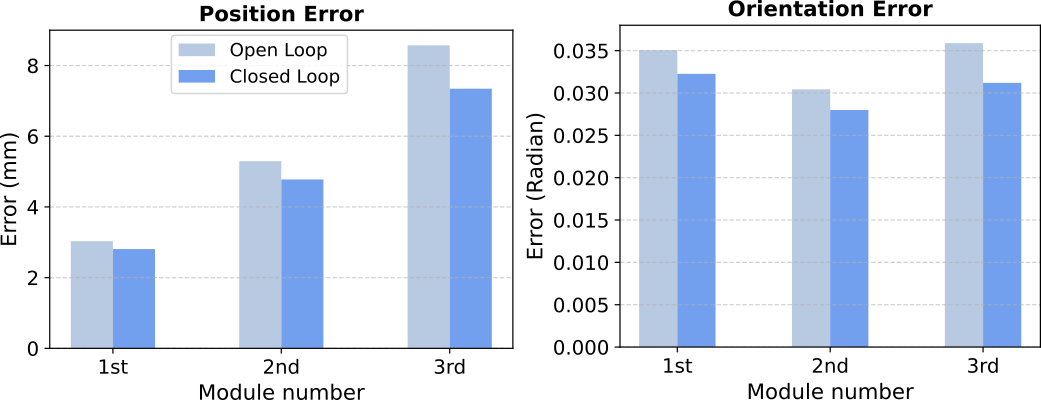}
	\caption{Left: Position error; Right: Orientation error for each module of the three-module soft robot (Exp2R) under open-loop and closed-loop settings. In the closed-loop case, pose error is fed back to the inverse model during both training and testing.}
	\label{fig:openclose_lat}
\end{figure}

\begin{table*}[]
\caption{Pose error of the SMPL and benchmark methods for the Exp2S experiment with a fixed number of modules. Errors are averaged across the three modules for each trajectory. All methods are trained and evaluated in a closed-loop setting.}
\label{sim_robo_exp2S}
\begin{center}
\begin{tabular}{|c|c|c|c|c|c|c|}
\hline
\textbf{Trajectory Type} & \textbf{Metrics} & \textbf{MLP} & \textbf{LSTM} & \textbf{VAE-LSTM} & \textbf{Bi-LSTM} & \textbf{SMPL} \\[1mm] 
\hline
\multirow{2}{*}{Test babbling} & Position (mm)& 7.942$_{\pm3.35}$ & 3.99$_{\pm2.24}$ & 3.97$_{\pm2.89}$ & 3.92$_{\pm3.32}$ & \textbf{3.52}$_{\pm2.99}$\\
& Orientation (degree)&  12.30$_{\pm{7.28}}$ & 7.02$_{\pm4.98}$ & 5.90$_{\pm3.64}$ & 5.63$_{\pm2.87}$ & \textbf{4.69}$_{\pm2.31}$\\
\hline
\multirow{2}{*}{Circle} & Position (mm) & 5.16$_{\pm3.77}$ & 2.49$_{2.15}$ & 2.34$_{1.94}$ & 2.38$_{\pm1.69}$ & \textbf{2.29}$_{\pm1.42}$ \\
& Orientation (degree) & 4.37$_{\pm2.49}$ & 2.03$_{\pm1.90}$ & 2.29$_{\pm1.87}$ & 1.99$_\pm{1.02}$ & \textbf{1.93}$_\pm{0.92}$ \\
\hline
\multirow{2}{*}{Rectangle} & Position (mm) & 1.71$_{\pm1.59}$ & 1.57$_{\pm1.27}$ & 1.29$_{\pm1.03}$ & 1.47$_{\pm1.11}$ & \textbf{1.27}$_{\pm0.70}$ \\
& Orientation (degree) & 1.61$_{\pm1.50}$ & 1.57$_{\pm1.22}$ & 1.35$_{\pm0.98}$ & 1.26$_{\pm1.01}$ & \textbf{1.24}$_\pm0.99$ \\
\hline
\multirow{2}{*}{Spiral} & Position (mm) & 2.91$_{\pm2.03}$ & 2.73$_{\pm1.93}$ & 1.57$_{\pm0.99}$ & 1.29$_{\pm0.96}$ & \textbf{1.24}$_{\pm0.85}$ \\
& Orientation (degree) & 3.09$_{\pm2.34}$ & 2.82$_{\pm1.76}$ & 1.68$_{\pm0.98}$ & \textbf{1.29}$_{\pm0.95}$ & 1.39$_{\pm0.83}$ \\
\hline
\end{tabular}
\end{center}
\end{table*}

\begin{table*}[]
\caption{Pose error of the fixed-size MSR for the real robot experiment (Exp2R) across multiple trajectories. Errors are averaged over the three modules, with all methods trained and evaluated in a closed-loop setting.}
\label{real_robo_exp2R}
\begin{center}
\begin{tabular}{|c|c|c|c|c|c|c|}
\hline
\textbf{Trajectory Type} & \textbf{Metrics} & \textbf{MLP} & \textbf{LSTM} & \textbf{VAE-LSTM} & \textbf{Bi-LSTM} & \textbf{SMPL} \\[1mm] 
\hline
\multirow{2}{*}{Test babbling} & Position (mm)& 15.98$_{\pm8.12}$ & 11.39$_{\pm 7.39}$ &9.99$_{\pm 7.05}$ &7.49$_{\pm 5.52}$ & \textbf{4.97}$_{\pm \textbf{3.73}}$\\
& Orientation (degree)&  6.10$_{\pm5.34}$ & 3.52$_{\pm2.96}$ & 3.30$_{\pm 3.12}$ & 2.47$_{\pm 1.83}$ &\textbf{1.75}$_{\pm \textbf{1.37}}$\\
\hline
\multirow{2}{*}{Circle} & Position (mm) &  13.06$_{\pm 10.84}$ & 10.97$_{\pm 8.11}$ & 4.18$_{\pm 3.83}$ & 2.94$_{\pm 2.60}$ & \textbf{2.77}$_{\pm \textbf{2.74}}$\\
& Orientation (degree) &  16.35$_{\pm 4.32}$& 14.95$_{\pm 3.53}$ & 4.33$_{\pm 2.97}$ & 2.0$_{\pm 1.44}$ & \textbf{1.87}$_{\pm \textbf{1.57}}$\\
\hline
\multirow{2}{*}{Rectangle} & Position (mm) & 5.49$_{\pm 5.42}$ & 4.73$_{\pm 3.59}$ & 3.34$_{\pm 2.99}$ & 2.59$_{\pm 2.48}$ & \textbf{2.38}$_{\pm \textbf{2.25}}$\\
& Orientation (degree) & 10.97$_{\pm 8.97}$& 5.29$_{\pm 4.74}$ & 4.60$_{\pm 3.80}$ & 4.14$_{\pm 3.22}$ & \textbf{3.69}$_{\pm \textbf{2.80}}$\\
\hline
\multirow{2}{*}{Spiral} & Position (mm) & 6.78$_{\pm 6.24}$ & 7.19$_{\pm 5.01}$ & 3.73$_{\pm 3.13}$ & 3.33$_{\pm 2.79}$ & \textbf{3.02}$_{\pm \textbf{2.29}}$\\
& Orientation (degree) & 16.98$_{\pm 6.56}$ & 15.32$_{\pm 7.14}$ & 11.18$_{\pm 4.25}$ & 8.48$_{\pm 3.98}$ & \textbf{5.66}$_{\pm \textbf{3.51}}$\\
\hline
\end{tabular}
\end{center}
\end{table*}

\begin{figure*}
	\centering
	\includegraphics[]{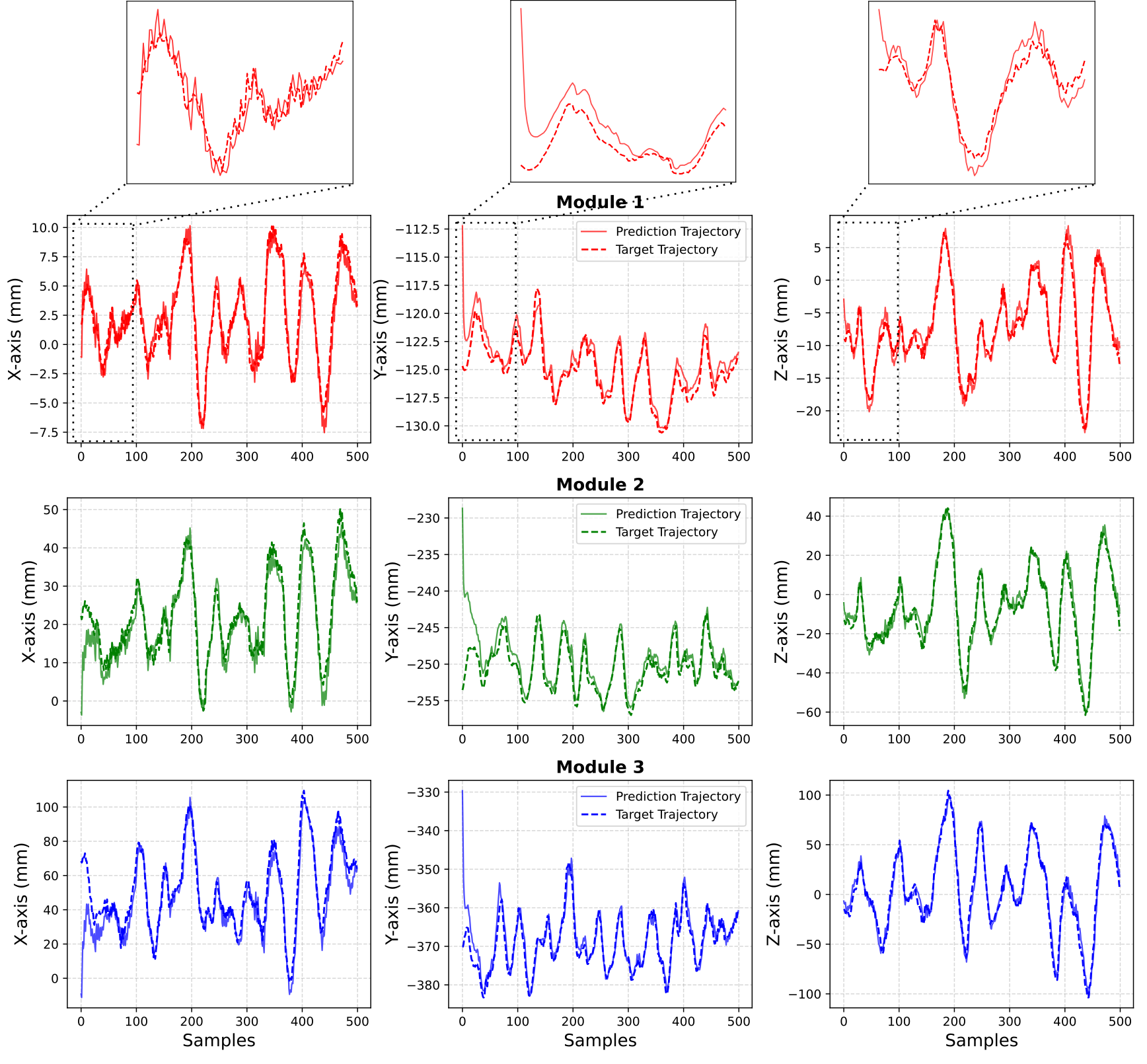}
	\caption{X, Y, Z position trajectories of the three-module soft robot (Exp2R) on the babbling test dataset. Red, green, and blue denote the trajectories of the first, second, and third modules along the x, y, and z axes, respectively.}
	\label{fig:babb_test_pred_act}
\end{figure*}

\begin{figure*}
	\centering
	\includegraphics[]{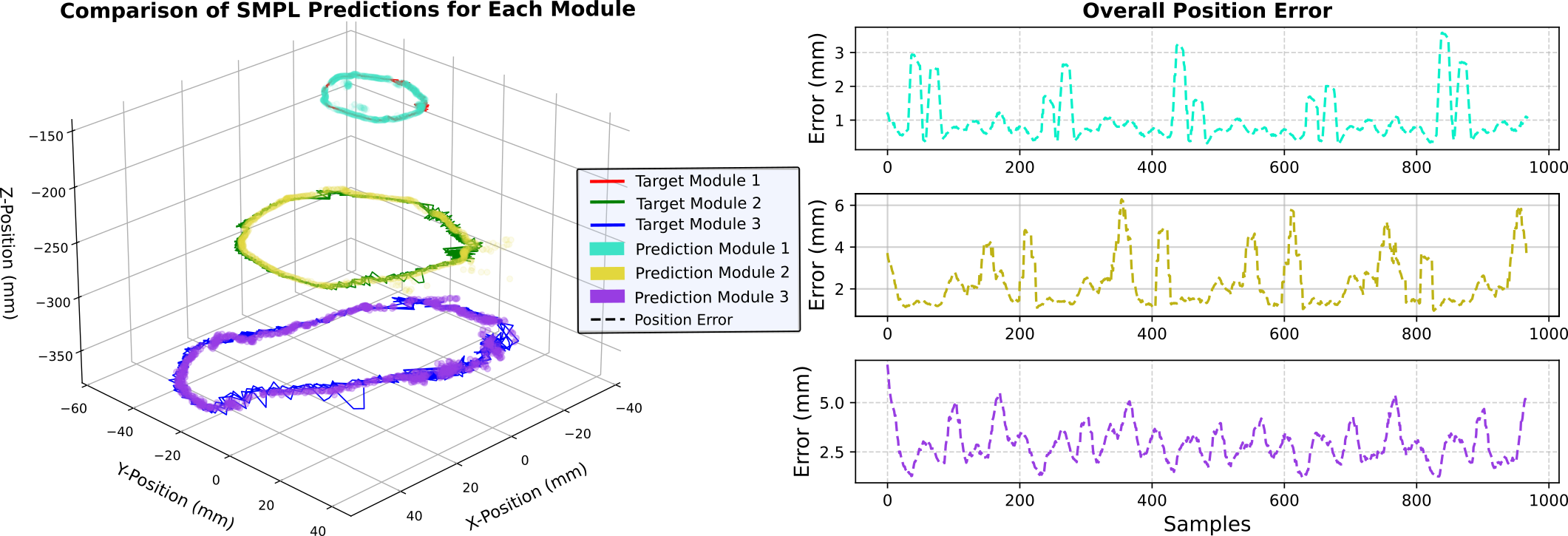}
	\caption{Left: The trajectory traced by each module of the MSR (Exp2R) to generate a rectangular shape. The solid lines depicts the target trajectory and the dispersed circles depicts the SMPL predictions. Right: Corresponding position error for each module, plotted as a dotted line and averaged across the three axes. The solid line depicts the predictions while the dashed line refer to the target trajectory.}
	\label{fig:circle_real_pred}
\end{figure*}

Table \ref{tab_exp1s_error} summarizes the position and orientation errors across different trajectories after sequential training on all five MSR configurations. The results show that the SMPL framework maintains high precision across all configurations, even after incremental learning. Furthermore, the table highlights the advantage of closed-loop training over open-loop control, consistently yielding lower errors across all configurations and trajectories. This improvement is further confirmed in Figure \ref{fig:openclose_lat} for the Exp2R experiment, where reduced pose error is observed for all three modules under the closed-loop setting.

\begin{figure*}
	\centering
	\includegraphics[]{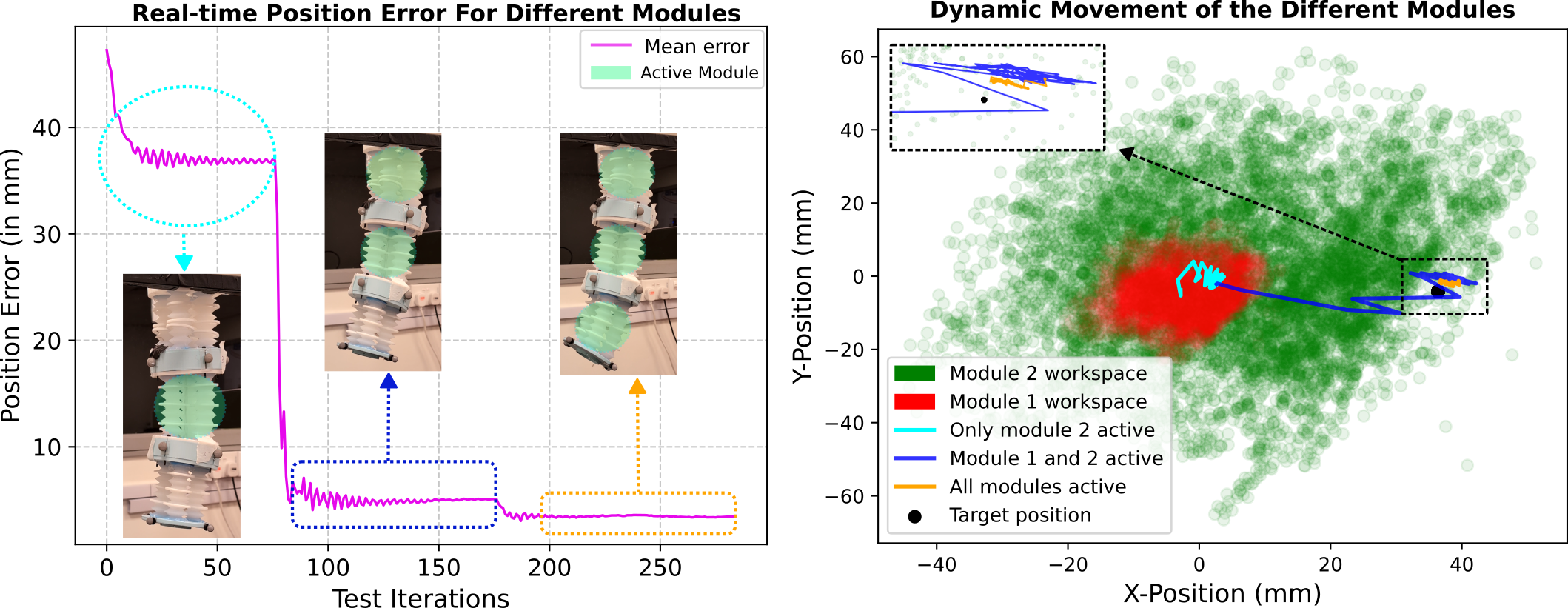}
	\caption{Left: Position error during sequential activation of the respective modules for reaching a virtual point in space. The algorithm dynamically activates the desired modules to minimize the position error. Right: The plot highlights the corresponding trajectory traced when activating the different modules to reach the desired virtual point (shown in black).}
	\label{fig:virtual_pt_exp}
\end{figure*}

We further assess the performance of the proposed approach in the Exp2S and Exp2R experiments, first in simulation and subsequently on the real robot. The simulation results, reported in Table \ref{sim_robo_exp2S}, demonstrate that the proposed method achieves the lowest error among all compared approaches. It is followed by the Bi-LSTM, the VAE-LSTM, and the standard LSTM, while the MLP baseline exhibits the highest error due to its limited capacity to model temporal dependencies. The superior performance of the proposed method can be attributed to its distributed architecture, where each module is controlled by a dedicated sub-network, enabling effective exploitation of local dynamics and improved overall accuracy.

Table \ref{real_robo_exp2R} presents the results obtained from real-world experiments. Consistent with simulation findings, the proposed method achieves the lowest position and orientation errors, followed by the Bi-LSTM, VAE-LSTM, standard LSTM, and the MLP baseline across all evaluated trajectories. Figure \ref{fig:babb_test_pred_act} illustrates the predicted (solid lines) and target (dashed lines) position trajectories along the X, Y, and Z axes for all three modules, represented in red, green, and blue, respectively. These results are obtained by directly applying the actuation predicted by the SMPL algorithm to the robot. The predicted trajectories closely follow the targets with high accuracy, except for minor deviations at the initial time step. This behavior arises from the initialization of previous actuation and pose error to zero, which are subsequently updated as the model begins generating predictions, as described in Section \ref{SMPL_training_labe}.

Figure \ref{fig:circle_real_pred} (left) shows a 3D visualization of the rectangular trajectory traced by each module of the MSR. The SMPL framework demonstrates good tracking accuracy; however, slight deviations from an ideal rectangular shape are observed for both the target as well as the predictions due to inherent structural constraints and asymmetric actuator distribution in the physical robot. Figure \ref{fig:circle_real_pred} (right) presents the corresponding position error for all three modules, averaged across the three spatial axes. The overall position error remains below 6 mm, with occasional peaks attributed to mechanical vibrations during motion.

\subsection{Adaptive Module Activation Experiment}
To assess the adaptive and distributed properties of the proposed SMPL algorithm, we design an experiment in which the second module of the MSR is required to reach a designated virtual point in space with minimal error by selectively activating only the necessary modules while keeping the other modules passive. A virtual target is defined within the workspace of the second module, as illustrated in Figure \ref{fig:virtual_pt_exp} (right). The objective is to reach this point while engaging the fewest possible modules. For this experiment, the orientation is held fixed at its initial value, and only the position space is controlled. The SMPL framework proceeds as follows:
\begin{enumerate}
\item Initially, only the second module is activated by sending the output of the corresponding SMPL sub-network, while the remaining modules are kept passive by applying zero pressure to the control inputs. The system attempts to reach the target using solely the second module, as shown in Figure \ref{fig:virtual_pt_exp} (left, cyan).
\item If the resulting position error still exceeds a predefined threshold, the preceding module is activated by incorporating the corresponding sub-network outputs, thereby augmenting the control input to the MSR.
\item The framework further assesses whether activating the subsequent module (i.e., the third) improves the reaching accuracy. If activation increases the error, the module is deactivated. In this experiment, incorporating the last module further reduced the overall position error, and it was therefore kept active (as shown in Figure \ref{fig:virtual_pt_exp}, yellow).
\end{enumerate}
After each activation, the algorithm introduces a short waiting period to allow the robot to stabilize. As shown in Figure \ref{fig:virtual_pt_exp} (right), when only the second module (cyan) is active, the reach error remains relatively high. Activating the preceding module substantially improves performance, allowing the robot to approach the target point. Activating the subsequent module (yellow) yields an additional, though smaller, reduction in error. Such selective module engagement is particularly advantageous in scenarios with limited onboard computational or power resources, as it enables more efficient and responsive control by activating only most relevant pressure valves necessary for that task.
\section{Conclusion}
\label{conclusion}
In this work, we implement a CL-based control framework capable of learning new MSR configurations incrementally and a distributed controller that can learn to control each module locally in case of an MSR with fixed number of modules. The proposed SMPL framework is evaluated on two simulation experiments as well as a real robot experiment for pose control. The algorithm is assessed on multiple trajectory tracking tasks and compared against standard benchmark algorithms from the literature. Our approach highlights improved performance compared to the benchmark methods. Furthermore, we also demonstrated the distributed control capabilities of our proposed approach through a dynamic reaching experiment, where the algorithm selectively activates only the necessary modules to reach a desired position, thereby reducing energy consumption and computational load.

However, a key limitation of the current approach lies in the increased training time due to the sequential learning of individual modules. As part of future work, we plan to integrate meta-learning principles with CL to enable faster adaptation to new modules with fewer training iterations. Additionally, we aim to explore the framework’s robustness under real-world conditions, particularly when actuators experience degradation or leakage due to aging, by selectively retraining the affected sub-network to restore system performance without full retraining. Also, we intend to extend the framework toward actuator-level modular control, assigning each actuator to its own sub-network to enable finer control and further improve the trade-off between energy efficiency and performance.
\bibliographystyle{IEEEtran}
\bibliography{references_file}

\begin{IEEEbiography}[{\includegraphics[width=1in,height=1.25in,clip,keepaspectratio]{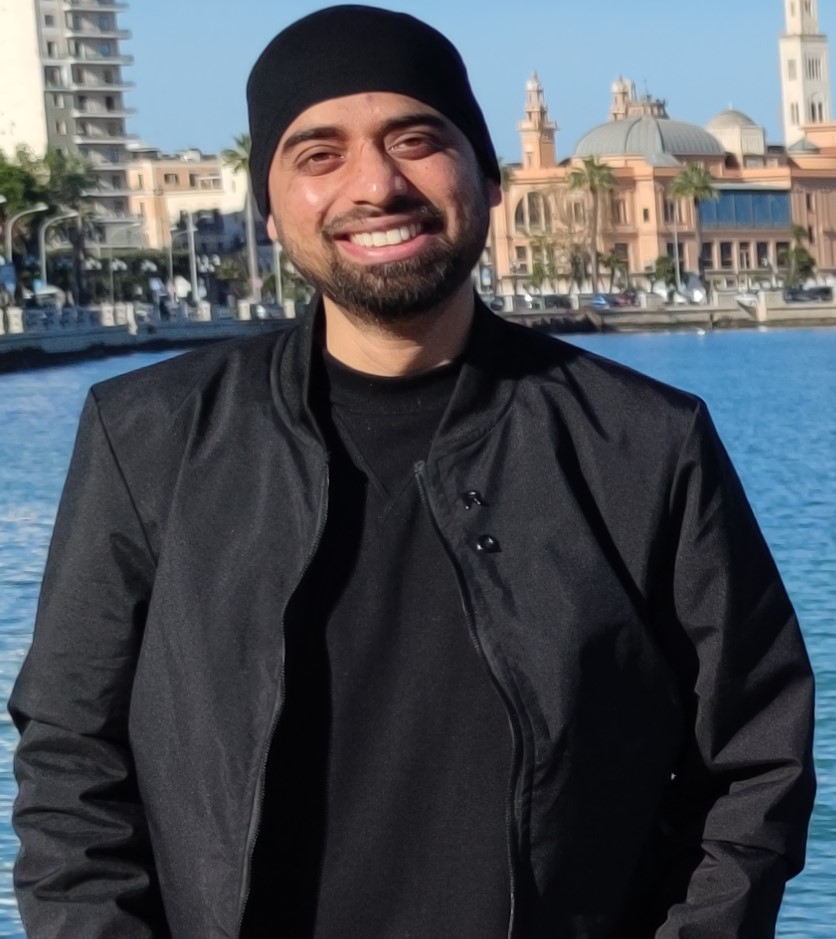}}]{Nilay Kushawaha} received the bachelor’s degree in
physics from the University of Delhi, India in 2020 and the master's degree in physics from Indian Institute of Technology Indore, India in 2022. He is currently pursuing PhD in Biorobotics and AI with The Biorobotics Institute, Scuola Superiore Sant’Anna, Pisa, Italy. His current research interests include continual learning, brain-inspired algorithms, reinforcement learning, and imitation learning.
\end{IEEEbiography}
\begin{IEEEbiography}[{\includegraphics[width=1in,height=1.25in,clip,keepaspectratio]{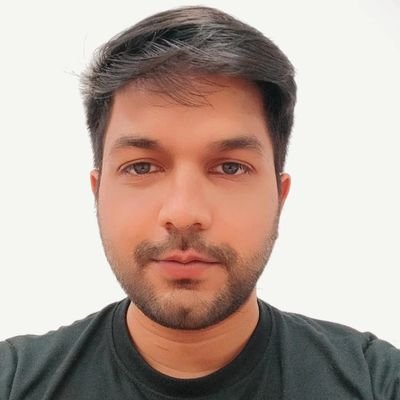}}]{Muhammad Sunny Nazeer} received the dual master’s degree in “European Master’s in Advanced Robotics Plus” under the Erasmus Mundus Joint Master’s Degree program from Ecole Centrale de Nantes, Nantes, France, in August 2020, and the Ph.D. degree in biorobotics under the Marie Skłodowska-Curie Actions (MSCA) SMART Innovative Training Network from the Department of Excellence in Robotics and AI, BioRobotics Institute, Scuola Superiore Sant’Anna, Pisa, Italy, in March 2024. Since April 2024, he has been a Postdoctoral Research Fellow with the Soft Robotics Lab, National University of Singapore, Singapore. His current research interests include imitative and adaptive control of self-healing and biodegradable soft robots, and leveraging advanced machine learning paradigms, such as imitation learning, reinforcement learning, and continual reinforcement learning.
\end{IEEEbiography}
\begin{IEEEbiography}[{\includegraphics[width=1in,height=1.25in,clip,keepaspectratio]{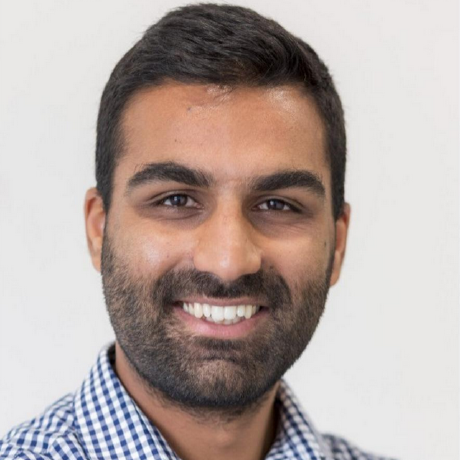}}]{Baljinder Singh Bal} is a PhD candidate at the CY Cergy Paris University and a Visiting scholar at the National University of Singapore, where his research focuses on the planning and control of assistive soft robots. He specializes in bio-inspired cognitive architectures and Active Inference, developing adaptive controllers for robots operating in contact-rich, unstructured environments. Prior to his PhD, he worked as a Robotics Software Engineer at the Italian Institute of Technology (IIT), developing real-time perception and control systems for heavy industrial manipulators. He holds a Master’s in Automation Engineering from the University of Bologna.
\end{IEEEbiography}
\begin{IEEEbiography}[{\includegraphics[width=1in,height=1.25in,clip,keepaspectratio]{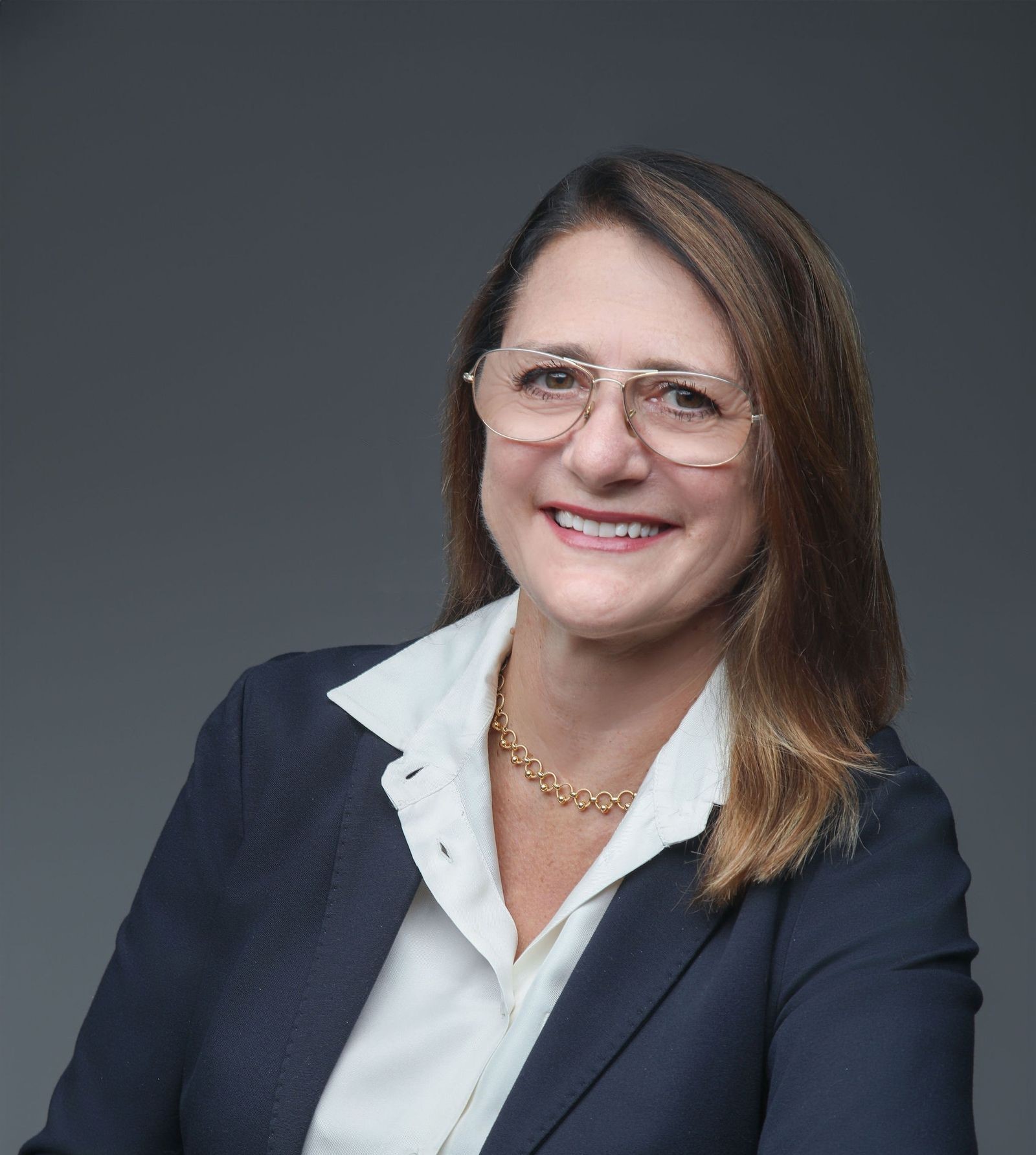}}]{Cecilia Laschi} received the degree in computer science from the University of Pisa, Pisa, Italy, the Ph.D. degree in robotics from the University of Genoa, Genoa, Italy, and the Honorary Doctorate degree from the University of Southern Denmark, Odense, Denmark, in 2023. She is currently the Provost’s Chair Professor of robotics with the National University of Singapore, Singapore, leading the Soft Robotics Laboratory. She is also the Director of the Advanced Robotics Centre. She was a JSPS Visiting Researcher with the Humanoid Robotics Institute, Waseda University, Tokyo, Japan. She is best-known for her research in soft robotics, an area that she pioneered and contributed to developing at an international level. She
investigates fundamental challenges for building robots with soft materials, with a bioinspired approach that started with a study of the octopus as a model for robotics. She explores marine applications of soft robots and their use in the biomedical field, with a focus on eldercare. She founded the IEEE International Conference on Soft Robotics (RoboSoft). She co-founded the spin-off company RoboTech. She is the Editor-in-Chief of Bioinspiration and Biomimetics, and member of the Scientific Advisory Board of Science Robotics.
\end{IEEEbiography}
\begin{IEEEbiography}[{\includegraphics[width=1in,height=1.25in,clip,keepaspectratio]{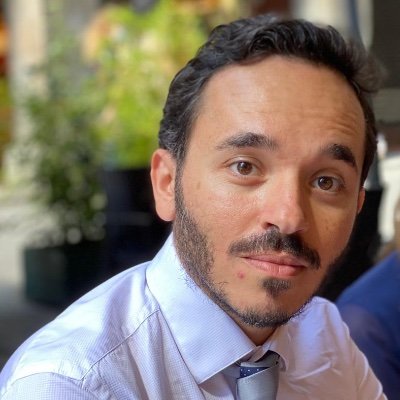}}]{Egidio Falotico} received the M.S. degree in computer science from the University of Pisa, Pisa, Italy, in 2008 and the Ph.D. degree in biorobotics from Scuola Superiore Sant’Anna (SSSA), Pisa, Italy, in 2013, and the Ph.D. degree in cognitive science from the University Pierre et Marie Curie, Paris, France, in March 2013. He is currently an Assistant Professor with The BioRobotics Institute, SSSA. He currently serves as the Deputy Leader and the Publications Manager in the Sub-Project 10 of the Human Brain Project. He is the author or coauthor of more than 40 international peer-reviewed papers and he regularly serves as a reviewer for more than 10 international ISI journals. He has been involved in some EU-funded projects (I-SUPPORT, SWARMs, SMARTE, RoboSoM, RobotCub), focusing on the development of brain-inspired algorithms for robot control. His research interests focus on neurorobotics, i.e., the
implementation of brain models from neuroscience in robots.
\end{IEEEbiography}

\EOD

\end{document}